\documentclass[sigconf]{acmart}
\usepackage{fvextra}

\usepackage{listings}
\usepackage{xcolor}

\AtBeginDocument{%
  }
    
\acmConference[ICAIL 2026]{International Conference on Artificial Intelligence and Law}{June 08-12}{Singapore}
\acmYear{2026}
\copyrightyear{2026}
 
\acmDOI{}
\acmISBN{}
\acmSubmissionID{}
\renewcommand\footnotetextcopyrightpermission[1]{}
\begin{document}
\title{Asking For An Old Friend: Diagnosing and Mitigating Temporal Failure Modes in LLM-based Statutory Question Answering}

\author{Max Prior}
\email{max.prior@tum.de}
\orcid{0009-0005-7066-996X}
\affiliation{%
  \institution{Technical University of Munich}
  \city{Garching near Munich}
  \country{Germany}
}

\author{Andreas Schultz}
\email{andreas.schultz@tum.de}
\orcid{0009-0002-2893-748X}
\affiliation{%
  \institution{Technical University of Munich}
  \city{Garching near Munich}
  \country{Germany}
}

\author{Matthias Grabmair}
\email{matthias.grabmair@tum.de}
\orcid{0000-0001-6586-2486}
\affiliation{%
  \institution{Technical University of Munich}
  \city{Garching near Munich}
  \country{Germany}
}



\begin{abstract}
Large language models are increasingly used for legal research, yet their fixed training cutoffs and reliance on static parametric knowledge are at odds with the evolving nature of statutory law. We study two temporal failure modes: \textit{post-cutoff staleness}, where models apply superseded rules after legislative amendments, and \textit{recency bias}, where models prefer newer provisions even when a historical version governs the fact pattern. To this end, we present a benchmark of 312 expert-validated, time-sensitive German statutory QA pairs spanning three categories: \textit{Post-Cutoff Amendment Questions}, \textit{Pre-Amendment Questions}, and \textit{Multi-Provision Pre-Amendment Questions}. We evaluate five LLMs by OpenAI, Anthropic and DeepSeek under four inference settings: \textit{Vanilla}, \textit{Web-search}, and two retrieval-augmented variants that enforce temporal validity via a fact date extraction and version filtering. Using an LLM-as-a-judge validated against human expert ratings, we find severe degradation in the \textit{Vanilla} post-cutoff setting. Both RAG approaches substantially improve performance across all question types, while web search yields unstable gains and exhibits a marked recency bias on historically anchored tasks. Our results indicate that reliable legal QA requires treating temporal validity as a hard constraint.
\end{abstract}

\maketitle

\section{Introduction}
The advent of large language models (LLMs) has led to substantial changes in everyday professional work, including in the legal sector. The Thomson Reuters report on generative AI from 2025 \cite{thomsonreuters2025genai_profservices} shows that the use of LLM-powered tools in legal practice increased from 14 \% in 2024 to 26 \% in 2025. Among current users, 40 \% report daily use, with legal research (73 \%) being among the most common applications. The survey of 1,702 respondents is dominated by common-law jurisdictions, with civil-law systems underrepresented. Notably, there were no respondents from Germany. To address this gap,
we draw on a survey conducted by the Bayerischer Anwaltverband (BAV). In 2024, the BAV surveyed 558 attorneys-at-law practicing in Bavaria \cite{bav2024ki_umfrage}. Among the attorneys who reported planned or actual use cases for AI tools, legal research emerged as the most frequently named area, with 73.1\% of respondents identifying it as a relevant application.

Because modern legal research is regularly performed using LLM-based systems, these findings 
motivate our work. Legal professionals may rely on LLMs without being aware of their limited temporal robustness.  
Unlike general knowledge, the correctness of a legal answer depends on the specific version of the provision and its temporal applicability. By contrast, LLMs are trained on static data and deployed with a fixed knowledge cutoff. This mismatch leads to two temporal failure modes. \textit{Post-cutoff staleness} occurs when a model relies on superseded provisions following legislative amendments. \textit{Recency bias} arises when, even though the historically applicable provision is available, the model incorrectly prioritizes newer legal texts.

\begin{figure}
    \centering
    \includegraphics[width=1\linewidth]{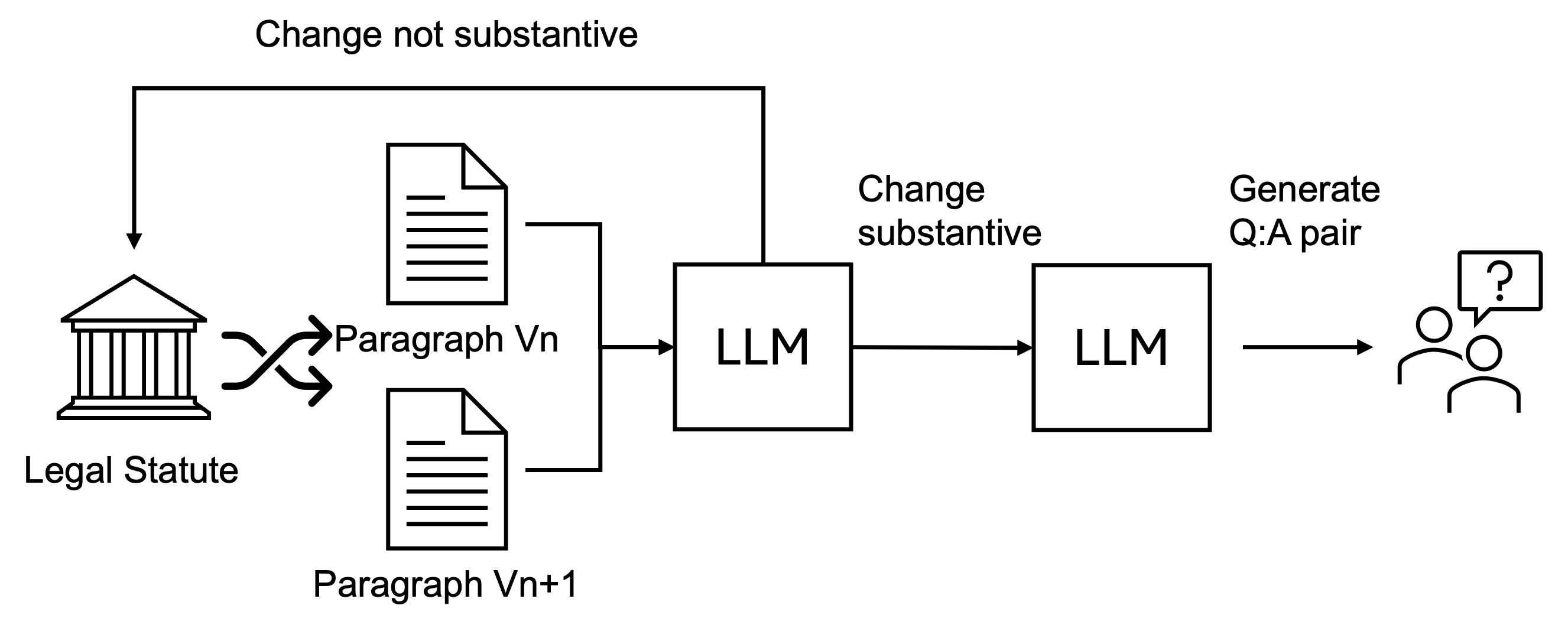}
    \caption{Overview of Pipeline for Post-Cutoff Amendment Question Generation}
    \label{fig:pipeline1a2}
\end{figure}

\section*{Contributions}
The contributions of this work can be summarized as follows:

\begin{itemize}   
    \item \textbf{Temporal Legal Benchmark:} We introduce a novel dataset of 312 expert-validated QA pairs designed to diagnose two temporal failure modes, specifically \textit{Post-Cutoff Amendment Questions} to identify staleness and \textit{Pre-Amendment Questions} to test an LLM's ability to apply historical versions of provisions.\footnote{The dataset is published under \url{https://github.com/TUMLegalTech/temporal_bias}.}
    
    \item \textbf{Diagnostic Evaluation:} We conduct an evaluation across multiple model tiers to isolate specific failure modes. For the \textit{Post-Cutoff Amendment} tasks, we evaluate \textbf{ChatGPT-5.1} and \textbf{Claude Sonnet 3.7}. Since their knowledge cutoffs precede our evaluation date, they serve to diagnose parametric staleness. For tasks requiring historical reasoning (\textit{Pre-Amendment} and \textit{Multi-Provision Pre-Amendment} questions), we use the state-of-the-art at experiment-time \textbf{ChatGPT-5.2}, \textbf{Claude 4.5 Opus}, and \textbf{DeepSeek v3.2}. Results reveal that even the most capable models frequently fail to override recent parametric knowledge when the legal fact pattern dictates a superseded provision.
    
    \item \textbf{Temporal RAG Mitigation:} We propose and evaluate two Retrieval-Augmented Generation (RAG) strategies — RAG-kNN and RAG-ToC — that use ``as-of date'' filtering. We demonstrate that strictly constraining the space to the relevant legal timeframe is the most effective mitigation for temporal degradation.
\end{itemize}

\section{Background}
\label{sec:background}

\subsection{Temporal generalization of LLMs}
\label{sec:background_temporal_generalization}

The challenge of time in NLP has been formalized as temporal generalization. Early work by Lazaridou et al. demonstrated that language model performance degrades as the evaluation data moves further beyond the training window \cite{NEURIPS2021_f5bf0ba0}. They argue that time should be treated as a distinct factor of generalization; a model must be evaluated on future data that does not temporally overlap with its training set to measure its adaptability. Zhu et al. extend this line of work by introducing \textit{FreshBench}, a time-stamped and extensible evaluation framework built from arXiv papers, news articles, and Wikipedia edits \cite{chenghaozhu-etal-2025-llm}. They identify two systematic behaviors: \textit{nostalgia bias}, where models perform better on older data, and \textit{neophilia bias}, where models over-favor recent information. In addition, they introduce the term \textit{Temporal Degeneration} which refers to the decline in performance in the future after the model is released. Experiments across proprietary and open-source models show that although newer models are more capable, they often exhibit faster temporal degeneration on future-oriented tasks. Recent empirical work has documented such failures in the medical domain: Vladika et al. introduce \textit{MedRevQA} and \textit{MedChangeQA} to evaluate temporal robustness in the biomedical domain, where verdicts change as new evidence emerges \cite{vladika-etal-2025-facts}. Evaluating eight major LLMs (such as  ChatGPT-4o and Llama 3.3), they find that performance declines sharply on questions requiring up-to-date verdicts, with models often outputting outdated answers due to pre-training corpora saturated with older literature. 

\subsection{Temporal shifts and adaption in the legal domain}
As law texts change over time, LLMs must generalize across time. In practice, this means two things: (i) models should remain reliable as legal provisions change, and (ii) models should still support historic legal provisions when the task demands it.

ChronosLex extends temporal generalization to multi-label legal text classification and shows that performance depends strongly on temporal training alignment \cite{t-y-s-s-etal-2024-chronoslex}. Models trained on older legal periods degrade substantially when evaluated on later periods, because legal language and label-conditional patterns drift over time. This aligns with the broader observation of Lazaridou et al. that the training order matters \cite{NEURIPS2021_f5bf0ba0}: if learning is not temporally structured, newer data can overwrite older patterns and yield brittleness under temporal shift. ChronosLex further demonstrates that training procedures which respect chronology, e.g., incremental updates across time, can significantly improve future performance when paired with continual-learning mechanisms that limit forgetting \cite{t-y-s-s-etal-2024-chronoslex}. In other words, temporal robustness is not only a data issue; it is also a training protocol issue. 

Coming from the downstream perspective, LexTempus \cite{t-y-s-s-vuong-2025-lextempus} targets upstream adaptation of legal language models. A key motivation is that naively updating a single model on the newest legal data risks catastrophic forgetting of older legal regimes, even though those older regimes may remain necessary for historical case analysis, retrospective compliance checks, or longitudinal legal research. Instead of overwriting the full model, LexTempus freezes the backbone language model and trains a growing pool of lightweight adapter experts. New experts are added dynamically when statistical shifts in loss or perplexity indicate a change in the legal data distribution. At inference time, a non-parametric router selects the expert (or expert combination) most relevant to the query's temporal and semantic context. This design enables the system to reuse older expertise when analyzing historical cases, rather than defaulting to the newest (and potentially inapplicable) legal regime.

Despite improved temporal robustness, both ChronosLex and LexTempus remain reactive: they adapt only after new statutes or legal interpretations appear in the training stream. As a result, they cannot reliably reason about entirely novel legislation or unprecedented legal developments until such changes are observed in data and incorporated through updates of their parametric knowledge. This limitation is consequential in law, where the relevant regime may change suddenly and correctness depends on recognizing the new authority immediately.

Given that the knowledge stored in a language model’s parameters is fixed once training is complete, RAG has become the standard technique for incorporating fresh information without retraining the model \cite{NEURIPS2020_6b493230}. In the legal domain the law is continually evolving, and RAG enables models to retrieve statutes, gazettes, and other publicly available materials so that answers are grounded in current law \cite{10.1145/3709025.3712219}. A notable example is the \textit{Statutory Research Assistant} (STARA) \cite{surani2025aiscalinglegalreform}, which automates statutory surveys. STARA preprocesses each statutory code by splitting it into granular provisions and annotating each fragment with its hierarchical structure, defined terms, and cross‑references. It then employs a large language model to classify these provisions and extract structured information according to task‑specific rubrics. A keyword filter produces a shortlist of candidate provisions, and the model is prompted to determine which are relevant and to extract the required fields. Because STARA operates over retrieved statutory text rather than relying on the model’s internal parameters, updating the statutory corpus (e.g. when laws change) suffices to keep the system up to date, with no need to retrain the underlying model. 

\subsection{Challenges}
Assessing legal RAG requires more than standard factual evaluation because misapplying doctrine is a primary risk. Zheng et al. emphasize that legal AI must handle a constantly changing corpus and therefore augment models with retrievers \cite{10.1145/3709025.3712219}. Their benchmark datasets pair questions with gold passages and answers to evaluate both retrieval and reasoning; they show that higher recall alone does not guarantee improved question‑answering because models must reason over the retrieved passages.

Another challenge is that recency bias can arise at both the retrieval and reasoning stages of LLM‑based systems. A key finding is that recency can function as an implicit relevance signal for LLM-based rerankers. By injecting synthetic publication dates into candidate passages while keeping their content unchanged, Fang et al. show that documents labeled as more recent are systematically promoted in the ranking \cite{10.1145/3767695.3769493}. This promotion is strong enough to measurably shift the average publication year of the top-ranked results, indicating that the model treats temporal freshness as evidence of relevance even when no content differences justify such a preference. Zhang et al. find that long‑context LLMs exhibit era sensitivity: models are markedly more accurate on modern cases than on nineteenth‑century ones \cite{zhang2025llms_overruled_precedent}. Their temporal stratification reveals that accuracy declines as the distance between overruling and overruled decisions grows, suggesting that models rely on temporal heuristics rather than doctrinal reasoning. They also motivate the use of abstention mechanisms as listed in \cite{wen-etal-2025-know}: when confronted with queries that exceed their temporal knowledge, models might be safer refusing to answer, offering a reasonable alternative to returning outdated or inapplicable legal conclusions.

\subsection{Diagnosing post-cutoff staleness and recency bias}

Zhang et al.’s \textit{LLMs Overruled?} dataset \cite{zhang2025llms_overruled_precedent} was designed specifically to probe recency bias in common law systems, but no comparable dataset exists for civil‑law jurisdictions, and none target post‑cutoff staleness. Büttner and Habernal’s GerLayQA \cite{buttner-habernal-2024-answering} dataset focuses on helping laypeople with German civil‑law questions: it pairs $\approx$ 21k real forum questions with lawyer‑written answers grounded in cited code paragraphs. Their raw data spanned 2004–2023, however they filter out QA pairs where the cited statutes had been amended. Consequently, no recency bias can be tracked by asking questions that would require amended laws. Additionally, they do not ensure that post-cutoff questions specifically require knowledge of the difference between the pre-cutoff and post-cutoff versions of the same provision. Therefore, it is unclear whether questions are being answered using pre-cutoff parametric knowledge making it impossible to track post-cutoff staleness accurately. Fan et al.’s LEXAM benchmark comprises 4886 law‑exam questions stemming from 2016–2023 with accompanying reference answers and explicit reasoning guidance; they evaluate models via an ensemble \textit{LLM‑as‑a‑Judge} \cite{fan2025lexambenchmarkinglegalreasoning}. They further observe that some answers may have become outdated and propose metadata‑based updates. However, LEXAM, like GerLayQA, does not test whether LLMs exhibit recency bias or struggle with post‑cutoff knowledge gaps. In contrast, our approach explicitly diagnoses both recency bias and post‑cutoff staleness in a legal setting, filling the gap left by existing benchmarks.

\section{Data}\label{sec:data}
\subsection{Data Sources and Preparation}
To evaluate temporal generalization, we required historical, time-stamped consolidated versions of statutory law. As Germany does not provide a public, legally authoritative database of historical consolidated statutes in machine-readable form \cite{prior-etal-2025-risks}, we used \texttt{buzer.de}, a privately operated resource offering unofficial consolidated versions of German federal law along with their in-force dates \cite{buzer2026}. Using this as raw data, we constructed a dataset of 312 legal questions spanning three categories. All QA sets were constructed synthetically using Gemini 3 Pro Preview. The complete prompt templates are provided in our GitHub repository.\footnote{See Footnote 1.}
 
We decided to construct the QA sets synthetically by shuffling from the historical consolidated statutes and feeding them into different LLM-based pipelines. We derived the paragraphs from six different statutes spanning entirely different German legal subdomains: (i) the German Civil Code (Bürgerliches Gesetzbuch, BGB), (ii) the German Code of Criminal Procedure (Strafprozessordnung, StPO), (iii) the Fiscal Code of Germany (Abgabenordnung, AO), (iv) the German Income Tax Act (Einkommensteuergesetz, EStG), (v) the German Federal Building Code (Baugesetzbuch, BauGB) and (vi) the German Commercial Code (Handelsgesetzbuch, HGB). By following this approach, we anticipate avoiding the overrepresentation of textbook examples and landmark legal cases which are likely prominent in LLM training data, while ensuring equal distribution across statutory provisions throughout different time periods and legal subdomains.

\subsection{Post-Cutoff Amendment Questions}

The first category, which we name \textit{Post-Cutoff Amendment Questions}, includes questions designed to probe post-cutoff staleness and therefore requires post-cutoff information to answer correctly. Using raw data, we identified paragraphs updated after the target LLMs’ training cutoff and constructed questions based on differences between the pre- and post-cutoff versions of the same provision, ensuring that correct answers depend on knowledge of legislative changes. We constructed such questions using an LLM-based pipeline (see Figure \ref{fig:pipeline1a2}). The pipeline randomly selects a paragraph with a version that starts after the cutoff and pairs it with the immediately preceding version, then prompts the model to confirm that it contains at least one substantive provision change. A provision version change is considered substantive when it results in a change in rights or obligations, requirements, deadlines, thresholds, percentages, amounts, or jurisdictions rather than mere editorial edits. We repeat this process until a substantive change is found.  After a substantive provision version change is confirmed we ask the LLM to generate a case question whose fact date falls after the cutoff so that the correct answer depends on knowing the post-cutoff version.\\

Here is an illustrative instance of the synthesized \textit{Post-Cutoff Amendment Questions} set (translated from German):\\

\textbf{Question:}

\textit{On February 10, 2025, Tenant M receives an ordinary notice of termination from Landlord V. Since M is unable to find alternative housing, he sends an email to V on February 20, 2025, objecting to the termination and outlining grounds for hardship. V rejects the objection, arguing that it is formally invalid because it was not received as a letter signed by hand. Was M’s objection submitted in a formally valid manner?}\\

\textbf{Answer:}

\textit{M’s objection was submitted in a formally valid manner. Formal validity is governed by Section 574b, Paragraph 1, Sentence 1 BGB. In the version in effect at the time of the declaration (February 2025), this provision dictates that the tenant's objection to termination must be declared in text form. Unlike "written form" (which requires a handwritten signature), text form (see Section 126b BGB) merely requires a legible declaration on a durable medium in which the person making the declaration is named. An email satisfies these requirements. Since M declared the objection via email, he adhered to the legally prescribed form. V's objection is therefore unfounded.}\\

In the version applicable prior to January 1, 2025, Section 574b, paragraph 1, Sentence 1 BGB (German Civil Code) stipulated that the tenant must declare their objection to the termination in writing. V would have been right in this case, as a simple email would not have been sufficient. The correct answer to the question therefore depends crucially on knowledge of the legal situation after the defined post-cutoff date. 

\subsection{Pre-Amendment Questions}

The second category, which we call \textit{Pre-Amendment Questions}, consists of questions where the correct answer depends on an older version of the law rather than the current one. We apply the same construction strategy as in the first category but reverse the roles: using the current provision leads to an incorrect answer, while correctly applying the outdated version is required. To construct such questions, we randomly select a paragraph with multiple versions available and classify each version transition to determine whether it contains substantive provision changes. Once a suitable candidate is found, we ask the LLM to generate a QA pair requiring the earlier version by embedding a fact date within the older version's validity period. The question implicitly requires temporal reasoning through the fact pattern rather than explicitly asking about the application of historical law.\\

Here is an illustrative instance of the synthesized \textit{Pre-Amendment Questions} set (translated from German):\\

\textbf{\textit{Question:}}

\textit{On March 3, 2017, the motorist M is stopped by Police Commissioner P during a routine traffic check. Since M smells strongly of alcohol and is slurring his speech, there is a suspicion of drunk driving pursuant to Section 316 of the German Criminal Code (Strafgesetzbuch, StGB). In order to secure evidence regarding this suspicion, P immediately orders a blood sample to be drawn by a physician. At this time (Tuesday morning), an investigating judge at the competent Local Court (\textit{Amtsgericht}) is available and on duty. However, P deliberately refrains from obtaining a judicial order, as he is of the opinion that he generally has the authority to decide for himself in traffic offenses of this nature. Was the order to draw blood issued by Police Commissioner P lawful?}\\

\textbf{\textit{Answer:}}

\textit{The order to draw blood by Police Commissioner P could be lawful pursuant to Section 81a StPO (German Criminal Procedure Code). This requires that P had the authority to issue the order. Pursuant to Section 81a, Paragraph 1 StPO, the authority to order such measures generally lies with the judge (\textit{Richtervorbehalt} / judicial prerogative). The public prosecutor’s office and its investigators (here, P) are only authorized to do so if the success of the investigation is endangered by delay (\textit{Eilkompetenz} / exigent circumstances). Such a risk exists if a judicial order cannot be obtained in time without jeopardizing the purpose of the measure. However, according to the facts of the case, an investigating judge was available. Since P deliberately failed to obtain a judicial decision where there was no concrete risk of delay, he acted without the necessary emergency authority. A statutory exception to the requirement of a judicial order specifically for road traffic offenses (such as Section 316 StGB) did not yet exist at the time the measure was taken. Consequently, the order issued by P was unlawful due to a lack of jurisdiction.}\\

The version of Section 81a, Paragraph 2, Sentence 2 StPO currently in force allows the police to order [a blood sample] themselves in cases of certain traffic offenses (including Section 316 StGB), even without a judicial order and without the existence of exigent circumstances. However, this new version only came into effect on August 24, 2017. Had the case occurred under the new version, P's order would have been lawful. Simply applying the current version of Section 81a StPO would lead to a legally incorrect conclusion.

\subsection{Multi-Provision Pre-Amendment Questions}

\begin{figure}
    \centering
    \includegraphics[width=1\linewidth]{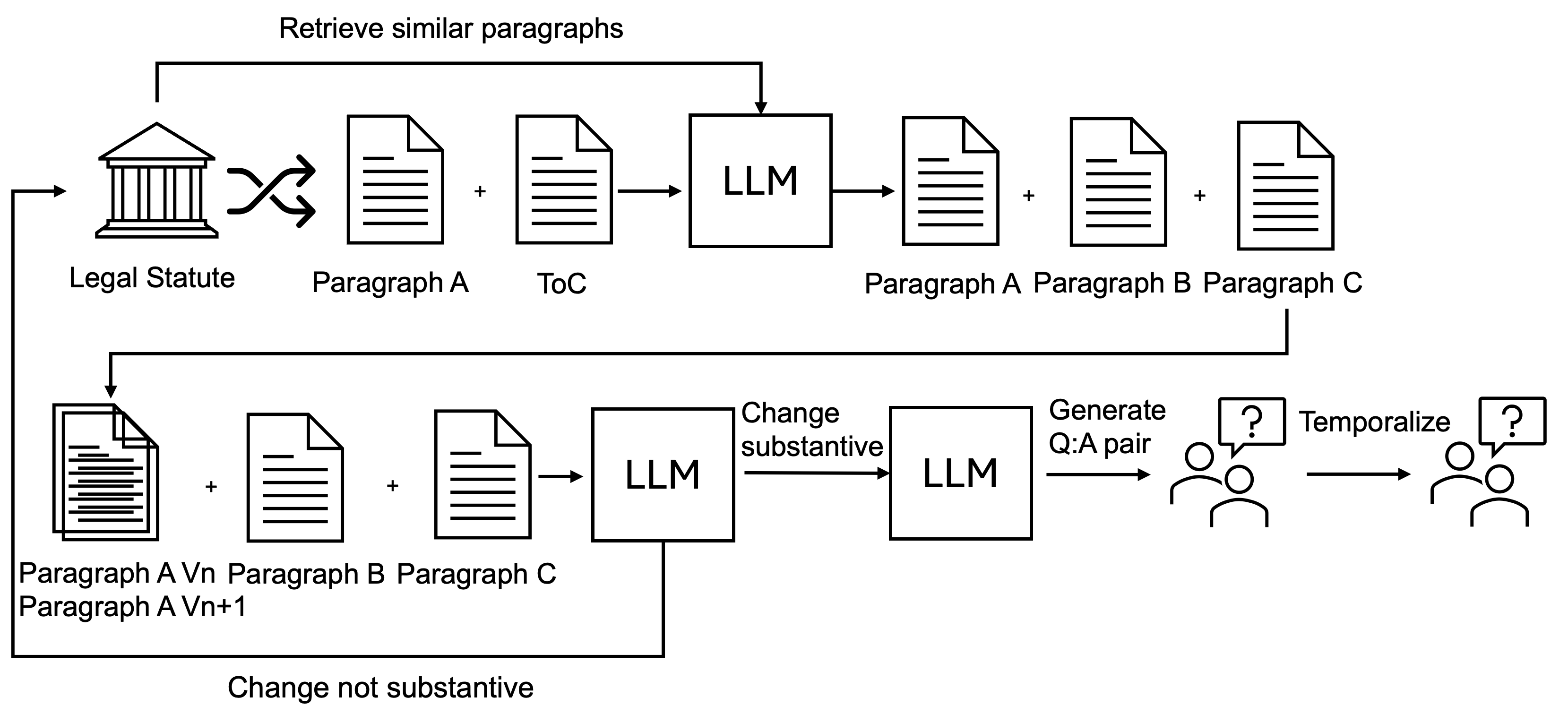}
    \caption{Overview of Pipeline for Multi-Provision Pre-Amendment Question Generation}
    \label{fig:pipeline3}
\end{figure}

The third category, termed \textit{Multi-Provision Pre-Amendment Questions}, extends the idea of Pre-Amendment Questions while providing more legally challenging QA pairs, as answering each question requires synthesizing information across multiple provisions rather than a single provision. We did not construct a post-cutoff multi-provision set analogue of the first category because the limited number of post–November 2024 version changes did not provide sufficient data. We constructed such questions using an LLM-based pipeline with three stages (see Figure \ref{fig:pipeline3}). First, we randomly select an anchor paragraph from a specific statute and present the statute's table of contents to the LLM, which we ask to select additional thematically connected paragraphs. At least one selected paragraph must have multiple historical versions, which is then validated using the same classification prompt to confirm substantive changes exist. Second, we ask the LLM to generate a temporally neutral case question based exclusively on the current versions of all selected paragraphs, requiring knowledge of each provision for a complete answer. Third, we ask the LLM to temporalize the draft question by injecting all historical versions of each paragraph and embedding a fact date that requires applying at least one provision in its earlier version while other provisions may remain current. 

Here is an illustrative instance of the synthesized \textit{Multi-Provision Pre-Amendment Questions} set (translated from German): 

\textbf{\textit{Question:}}

\textit{G-GmbH is a large corporation that has commissioned auditor P to audit its annual financial statements. P conducts the audit but negligently overlooks significant irregularities in the accounting that give a false picture of the company's financial position and should have been detected if the audit had been carried out conscientiously. On March 10, 2020, P summarizes the results of his audit in writing and declares in an audit opinion that the financial statements comply with legal requirements without making any reservations. As a result of this incorrect audit opinion, G-GmbH later suffers a financial loss of EUR 1.2 million. Is G-GmbH entitled to claim damages from P in the full amount (EUR 1.2 million) for the incorrect audit?}\\

\textbf{\textit{Answer (shortened):}}

\textit{G-GmbH is not entitled to claim damages from P in full, but only in the amount of EUR 1 million pursuant to Section 323, Paragraph 1, Sentence 3 HGB in conjunction with Section 323, Paragraph 2, Sentence 1 HGB (in the version applicable on March 10, 2020).}\\

\textit{I. Basis for the claim }\\
\textit{The basis for the claim for damages is Section 323, Paragraph 1, Sentence 3 HGB. [...]}\\

\textit{II. Applicability and duty to examine (Section 316 HGB)}
\textit{[...]}\\

\textit{III. Breach of duty (Section 317 HGB, Section 322 HGB)}\\
\textit{P breached his duties by negligently overlooking material irregularities (Section 317, Paragraph 1, Sentence 3 HGB) and issuing an unqualified audit opinion (Section 322, Paragraph 3 HGB), even though this should have been qualified or refused (Section 322, Paragraph 4 HGB).}\\

\textit{IV. Fault and damage}
\textit{P acted negligently. G-GmbH suffered a causal loss of EUR 1.2 million.}\\

\textit{V. Limitation of liability (Section 323, Paragraph 2 HGB)}
\textit{The claim is limited in amount. Pursuant to Section 323, Paragraph 2, Sentence 1 HGB in the version applicable at the time of the breach of duty in March 2020, the liability for damages in the event of negligent conduct is limited to one million euros for one audit. Since G-GmbH is not a listed stock corporation (cf. Section 323, Paragraph 2, Sentence 2 HGB, old version), no higher liability limit applies.}\\

\textit{Result: G-GmbH can claim damages from P, but limited to the maximum amount of EUR 1 million. There is no claim to the full amount of EUR 1.2 million.)}\\

Answering this legal question requires, inter alia, reviewing Section 323 HGB (German Commercial Code). The case occurs in March 2020, when the version of Section 323 HGB valid at that time capped auditor liability for negligence at 1 million euros. The current version of the same provision raised this limit to 1.5 million euros for non-capital-market-oriented companies. Under current law, the 1.2 million euro damage would be fully recoverable. The correct solution — limiting recovery to 1 million euros — requires applying the law in force at the time of the facts.

\subsection{Human Expert Evaluation of Synthesized QA sets}

While all QA pairs were fully LLM generated, each pair was subsequently evaluated by legal experts (two senior law students and one fully-qualified lawyer) to ensure quality control across each set. From a total of 697 generated pairs, 312 were accepted. The primary reasons for rejection were: (i) despite the classification prompt's instructions, the version change was only editorial in nature rather than substantive, (ii) the question implicitly consisted of multiple independent sub-questions rather than a single coherent legal issue, (iii) the legal consequence remained identical regardless of which version was applied, contradicting the temporal sensitivity requirement, and (iv) the generated answer contained legal errors. The accepted 312 questions form our final evaluation dataset, comprising 115 Post-Cutoff Amendment Questions, 113 Pre-Amendment Questions, and 84 Multi-Provision Pre-Amendment Questions. 39 of the accepted 312 questions constitute \textit{duplicate pairs}. Pairs are considered to be \textit{duplicate} if they are not identical in wording but reference the same legal provision and the same amendment event. Such pairs construct distinct fact patterns around a single legislative change, differing in scenario framing, parties involved, and specific fact dates. Although identified post-hoc, these pairs were deliberately retained as an internal consistency check, since they allow us to verify whether models produce stable scores across different fact patterns targeting the same legislative change.

\section{Experiments}
\subsection{Model and RAG Configurations}
For the \textit{Post-Cutoff Amendment Questions}, we set the knowledge cutoff to November 1, 2024. Models whose training data precede this cutoff include ChatGPT-5.1 and Claude Sonnet 3.7. We did not include a DeepSeek model since there is no publicly available, reliable source for the knowledge cutoff date. Correctly answering questions in this category therefore requires access to legislative updates introduced after this date, enabling an explicit evaluation of post-cutoff knowledge limitations.

For \textit{Pre-Amendment Questions} and \textit{Multi-Paragraph Pre-Amendment Questions}, we evaluated state-of-the-art models ChatGPT-5.2, Claude Opus 4.5, and DeepSeek V3.2. In these categories, the challenge is to determine whether models can recognize the need to apply historical versions of the law, even when the most recent paragraph versions are available. ChatGPT-5.1's older knowledge cutoff isolates post-cutoff staleness, whereas the newer ChatGPT-5.2 tests if a state-of-the-art model can override its current knowledge to apply historical provisions.

\begin{figure}
    \centering
    \includegraphics[width=1\linewidth]{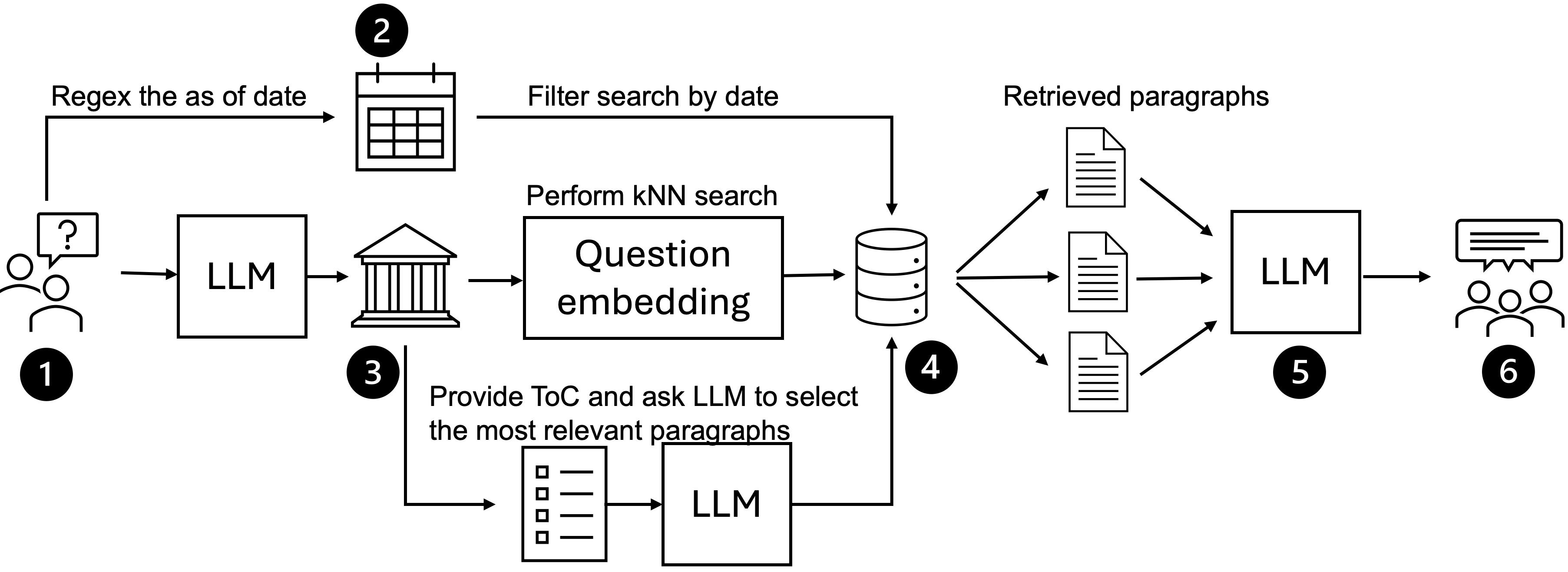}
    \caption{RAG-kNN and RAG-ToC}
    \label{fig:rag_knn_toc}
\end{figure}

We evaluated the models in four distinct experimental settings that differ in the information available at inference time:

\begin{itemize}

\item\textit{LLM (vanilla)} relies exclusively on the model's internal parametric knowledge, with no access to external tools or data. The prompt instructs the model to act as a German legal expert, answer precisely based solely on provided legal texts, and return a fixed refusal string when information is insufficient.\footnote{The full prompts for all experimental settings are provided in our GitHub repository, see Footnote 1.}
\mbox{}

\item\textit{LLM-web} enables models to answer using online sources, distinguishing between native tool-calling and external context injection. For ChatGPT-5.1, ChatGPT-5.2, Claude Sonnet 3.7 , and Claude Opus  4.5, we utilize their native, integrated web search tools via their respective APIs. In this setting, the model acts as an autonomous agent: it reasons whether external information is required and, if so, triggers a tool-call to formulate and execute its own search queries.

In contrast, for DeepSeek v3.2, which lacks a native web search interface, we implement a deterministic retrieval pipeline. We use Gemini-based grounding Google Web Search to fetch snippets through the Google Search API. These web snippets are then injected into the local OpenAI-compatible prompt. Unlike the autonomous models, DeepSeek v3.2 is a passive recipient of this data as the search is executed by external scripts before the model is even invoked. This forces the model to process the provided context regardless of its own internal reasoning.




\item \textit{RAG-kNN} follows the retrieval-augmented workflow illustrated in Figure \ref{fig:rag_knn_toc}. After a user submits a question (1), a regex pattern extracts the $as-of$ date from the query (2), acting as a hard constraint for legal validity by filtering the retrieval space to ensure only paragraph versions valid during the relevant period are considered. Simultaneously, the respective LLM identifies the specific applicable law (3).

To perform retrieval, the question is embedded using the \texttt{text-embedding-3-large} model via the OpenAI API. The system executes a $k$-nearest-neighbor search ($k=6$) utilizing cosine similarity over the embeddings of the filtered full-text paragraphs chunks (4). These top-ranked chunks serve as keys to retrieve their corresponding paragraphs. Finally, these paragraphs are provided as context to the LLM (5), ensuring complete legal nuance is preserved in the generated answer (6).

\item\textit{RAG-ToC} follows a different retrieval strategy while sharing the same initial law-identification and temporal filtering steps (1-3). Instead of embedding-based retrieval, the model is presented with the selected statute's table of contents, stored as a structured outline, and instructed to select a small set of relevant paragraphs (4). The full text of these paragraphs is then retrieved from the local database and supplied as context. The remaining steps are the same as for RAG-kNN.
\end{itemize}

We evaluate model answers following an LLM-as-a-judge approach, where a separate LLM scores each candidate answer against a fixed rubric and returns structured ratings. This approach follows prior work showing that rubric-guided LLM judges can align well with human evaluations while enabling scalable, open-ended assessment \cite{zheng2023judging, enguehard2025lemaj}. We used Gemini 3 Flash Preview as the judge LLM which we prompted to evaluate each QA pair against the respective candidate answer based on the following criteria: 

\begin{itemize}
    \item Outcome correctness: Is the legal conclusion made by candidate correct?
    \item Legal reasoning correctness: Is the candidate's derivation of the legal conclusion correct?
    \item Legal basis correctness: Did the candidate identify the correct provisions necessary to answer the question?
    \item Version application correctness: Did the candidate apply the right version of the identified legal provisions? 
\end{itemize}

Each criterion was scored on a continuous scale from 0 to 1. The complete prompt template is provided in our GitHub repository.\footnote{See Footnote 1.}\\

\subsection{Post-Cutoff Questions}

\begin{figure*}[t]
    \centering
    \includegraphics[width=1\linewidth]{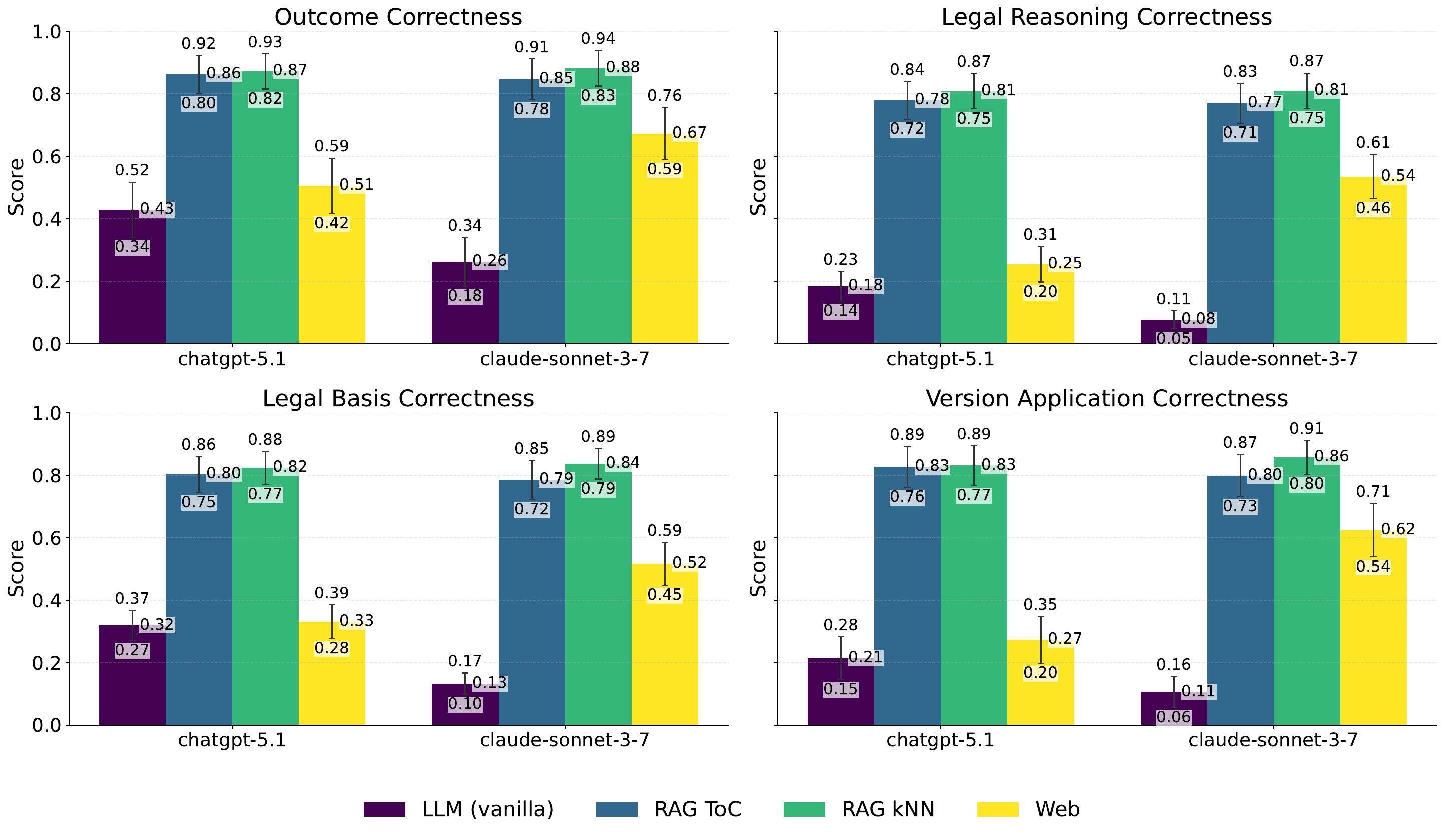}
    \caption{Post‑Cutoff Questions: Comparison of metrics across models and methods.}
    \label{fig:postcutoff-metrics}
\end{figure*}

In total, we evaluated 115 Post-Cutoff Questions. Figure \ref{fig:postcutoff-metrics} summarizes their evaluation with respect to the different models and methods. Both RAG methods achieve consistently high scores across all metrics (0.78--0.88). In contrast, the \textit{LLM (vanilla)} baseline shows severe degradation, particularly in reasoning and version application (Claude Sonnet 3.7: 0.08--0.11; ChatGPT 5.1: 0.18--0.21), confirming that post-cutoff questions expose substantial parametric staleness. Web search yields moderate scores (0.25--0.67) but remains less reliable than retrieval grounded in a curated law corpus.

Using Welch two-sample t-tests \cite{10.1093/biomet/34.1-2.28} on the post-cutoff summary statistics, all four metrics — outcome correctness, legal reasoning correctness, legal basis correctness, and version application correctness — shows significant mean differences between LLM (vanilla) and each RAG method, and between each RAG method and web search ($p < 0.05$) for both models. In contrast, the difference between RAG ToC and RAG kNN is not statistically significant for any metric.

\subsection{Pre-Amendment Questions}
\begin{figure*}[t]
    \centering
    \includegraphics[width=1\linewidth]{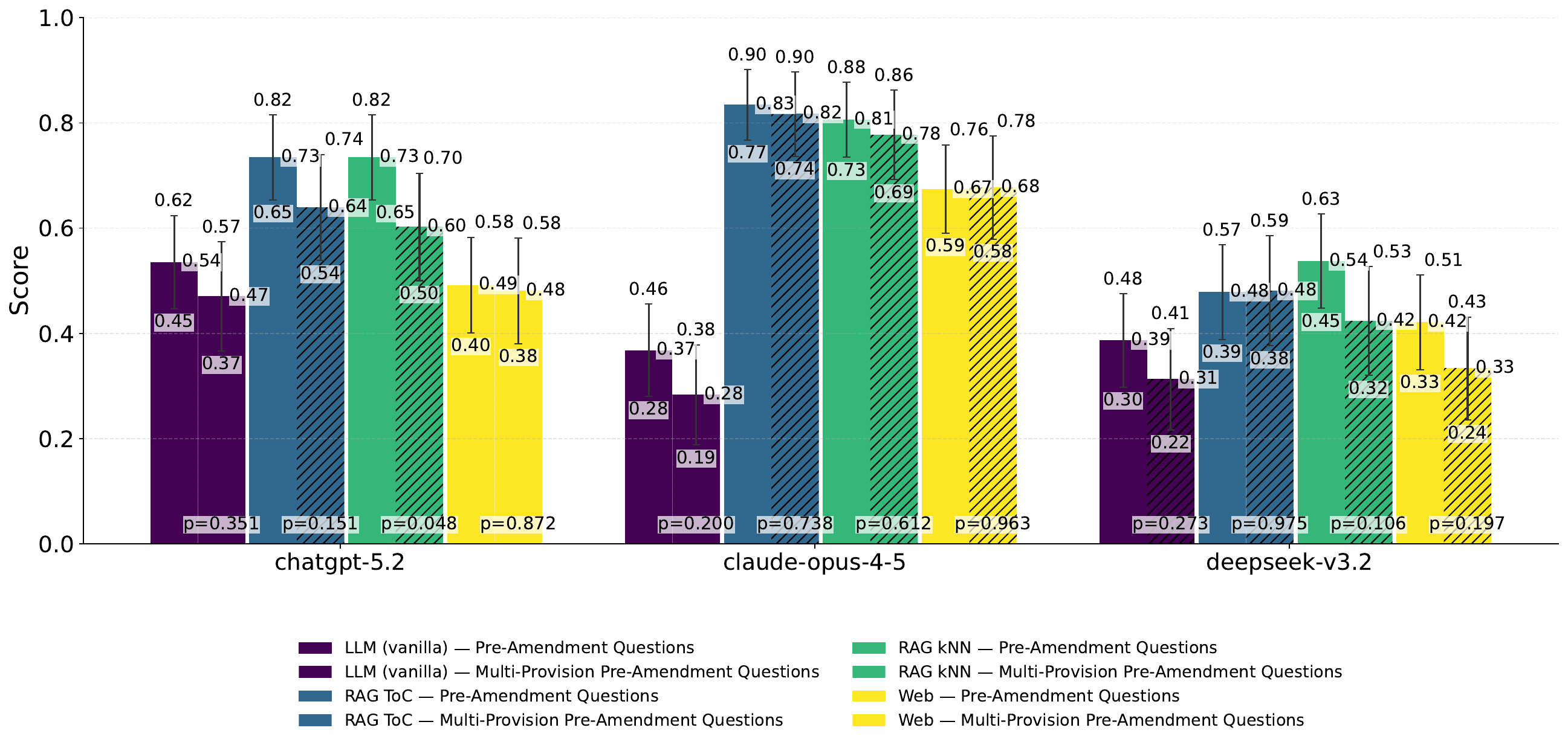}
    \caption{Pre-Amendment Questions: Single vs. Multi-Provision (Outcome Correctness).}
    \label{fig:single_vs_multi_outcome_mean}
\end{figure*}

We evaluated 113 single-provision and 84 multi-provision pre-amendment questions. Figure \ref{fig:single_vs_multi_outcome_mean} compares outcome correctness across ChatGPT-5.2, Claude Opus  4.5, and DeepSeek v3.2. On single-provision tasks, both RAG methods consistently outperform baselines, with RAG-ToC (0.48--0.83) and RAG-KNN (0.54--0.81) surpassing \textit{Vanilla} (0.37--0.54) and Web search (0.42--0.67).

Transitioning to multi-provision questions generally incurs a modest performance penalty. LLM (Vanilla) scores drop to 0.28--0.47 and RAG-KNN to 0.42--0.78, while RAG-ToC remains robust (0.48--0.82). We assessed the statistical significance of this degradation using Welch two-sample t-tests. Notably, only ChatGPT-5.2 with RAG-KNN showed a significant decline ($p=0.048$; $0.73 \rightarrow 0.60$). All other model-method pairs showed no statistically significant difference ($p > 0.11$), suggesting that while multi-provision tasks are more complex, they do not systematically break the models' reasoning capabilities when the correct legal context is retrieved.

\subsection{Perfect Score Analysis}

\begin{table*}[t] 
    \centering
    
    \footnotesize 
    \setlength{\tabcolsep}{2.5pt} 
    \renewcommand{\arraystretch}{1.2}
    
    \begin{tabular}{@{} ll *{4}{ccccc} @{}}
        \toprule
        \textbf{Model} & \textbf{Variant} & 
        \multicolumn{5}{c}{\textbf{LLM (vanilla)}} & 
        \multicolumn{5}{c}{\textbf{RAG ToC}} & 
        \multicolumn{5}{c}{\textbf{RAG kNN}} & 
        \multicolumn{5}{c}{\textbf{Web}} \\
        \cmidrule(lr){3-7} \cmidrule(lr){8-12} \cmidrule(lr){13-17} \cmidrule(lr){18-22}
        & & CO & R & LB & V & RC & CO & R & LB & V & RC & CO & R & LB & V & RC & CO & R & LB & V & RC \\
        \midrule
        ChatGPT-5.1 & Post-Cutoff & 40.00 & 0.00 & 3.48 & 13.91 & 0 & 83.48 & 46.09 & 47.83 & 73.04 & 2 & 83.48 & 49.57 & 46.96 & 74.78 & 1 & 46.96 & 3.48 & 3.48 & 16.52 & 0 \\
        Claude Sonnet 3.7 & Post-Cutoff & 24.35 & 0.00 & 0.00 & 5.22 & 22 & 83.48 & 49.57 & 47.83 & 71.30 & 2 & 86.96 & 51.30 & 49.57 & 70.43 & 0 & 63.48 & 17.39 & 19.13 & 55.65 & 0 \\
        \midrule
        ChatGPT-5.2 & Pre-Amendment & 49.56 & 12.39 & 14.16 & 34.51 & 1 & 72.57 & 47.79 & 45.13 & 47.79 & 15 & 72.57 & 45.13 & 51.33 & 55.75 & 11 & 46.90 & 25.66 & 37.17 & 45.13 & 0 \\
        ChatGPT-5.2 & Multi Pre-Amendment & 42.86 & 4.76 & 4.76 & 17.86 & 3 & 61.90 & 28.57 & 26.19 & 50.00 & 12 & 57.14 & 17.86 & 15.48 & 40.48 & 20 & 39.29 & 8.33 & 11.90 & 30.95 & 0 \\
        \midrule
        Claude Opus 4.5 & Pre-Amendment & 35.40 & 3.54 & 6.19 & 15.04 & 25 & 81.42 & 48.67 & 59.29 & 68.14 & 1 & 78.76 & 53.10 & 62.83 & 73.45 & 0 & 63.72 & 41.59 & 53.10 & 62.83 & 0 \\
        Claude Opus 4.5 & Multi Pre-Amendment & 26.19 & 0.00 & 2.38 & 9.52 & 11 & 78.57 & 36.90 & 34.52 & 66.67 & 0 & 73.81 & 28.57 & 20.24 & 52.38 & 0 & 64.29 & 32.14 & 29.76 & 58.33 & 0 \\
        \midrule
        DeepSeek v3.2 & Pre-Amendment & 37.17 & 3.54 & 4.42 & 12.39 & 19 & 44.25 & 25.66 & 38.05 & 46.90 & 1 & 50.44 & 33.63 & 42.48 & 47.79 & 2 & 39.82 & 5.31 & 8.85 & 15.04 & 1 \\
        DeepSeek v3.2 & Multi Pre-Amendment & 27.38 & 0.00 & 1.19 & 9.52 & 14 & 45.24 & 11.90 & 15.48 & 32.14 & 2 & 39.29 & 7.14 & 10.71 & 29.76 & 2 & 29.76 & 1.19 & 1.19 & 10.71 & 1 \\
        \bottomrule
    \end{tabular}
        \caption{\textbf{Correct Outcome (CO) \% | Reasoning (R) \% | Legal Basis (LB) \% | Version (V) \% | Refusal Count (RC)} \newline 
    Sample sizes: Post-Cutoff ($n=115$), Pre-Amendment Single ($n=113$), Multi ($n=84$).}
    \label{tab:detailed_results_percentage}
\end{table*}

Table \ref{tab:detailed_results_percentage}  presents the percentage of perfect scores across the four metrics. RAG methods consistently outperform other variants in Post-Cutoff scenarios, where \textit{LLM (vanilla)} baselines struggle with parametric staleness. For example, while Claude Sonnet 3.7 Vanilla achieves an outcome correctness of only 24.35\%, RAG kNN reaches 86.96\%. This Vanilla degradation is frequently accompanied by high refusal rates; Claude Sonnet 3.7 generated 22 one-line refusals, adhering to the explicit instruction to state \textit{The information provided is not sufficient to answer the question} when uncertain.

The results further underscore the increased complexity of Multi-Provision tasks relative to Pre-Amendment Single-Provision questions. Baselines exhibit a marked decrease in reasoning and legal basis correctness when multiple provisions are involved. For instance, the outcome correctness for Claude Opus 4.5 Vanilla drops from 35.40\% to 26.19\% when moving to Multi-Provision tasks, whereas RAG ToC maintains a more robust performance, shifting from 81.42\% to 78.57\%. Across all models, RAG methods provide substantial improvement in citing the correct legal basis and applying the appropriate temporal version of the law.

\subsection{Recency Bias}\label{subsec:recency_bias}

\begin{figure*}[t]
    \centering
    \includegraphics[width=1\linewidth]{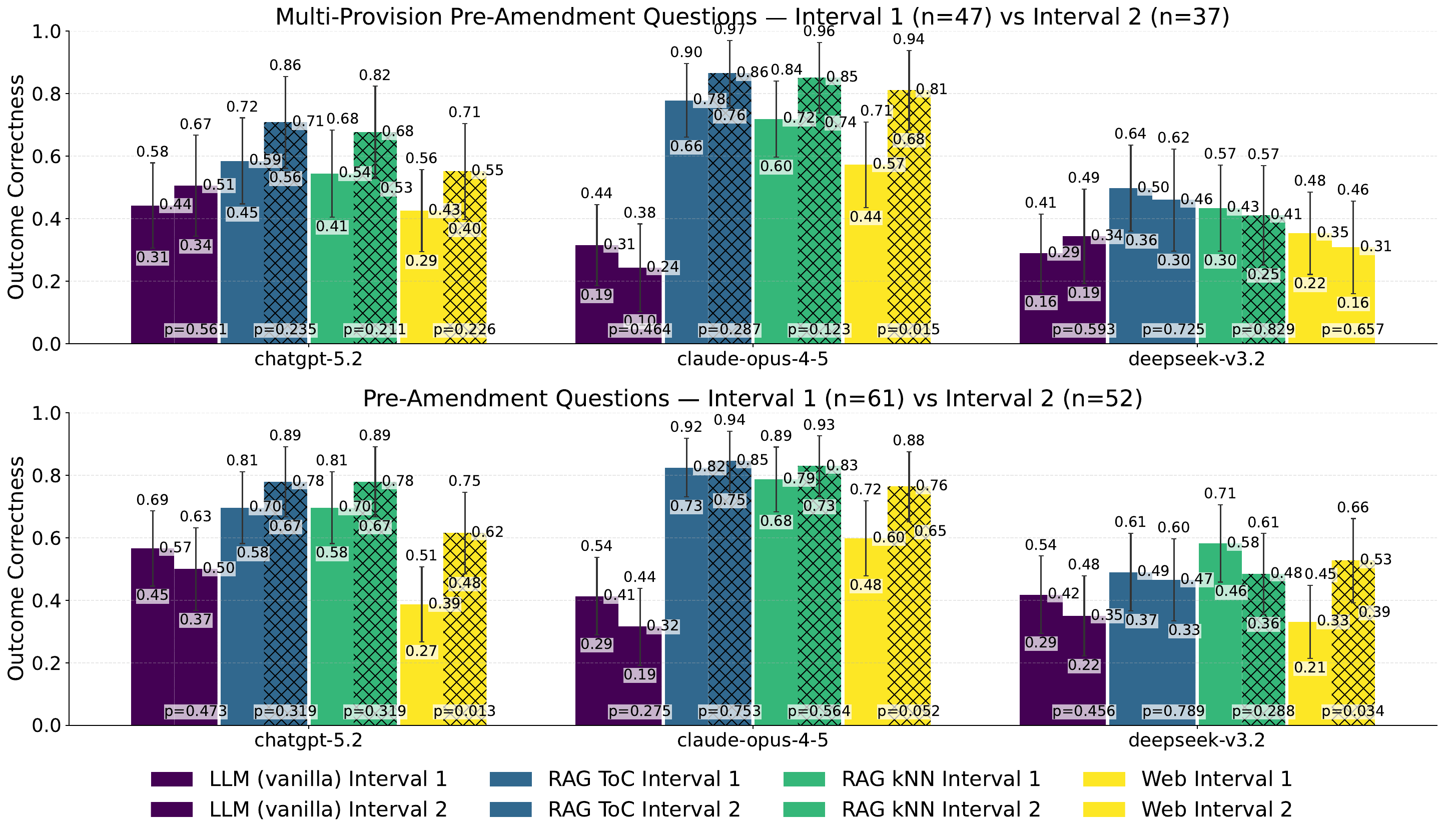}
    \caption{Recency Bias: Outcome Correctness split by time interval (Interval 1: Older, Interval 2: Recent).}
    \label{fig:recency_outcome_correctness}
\end{figure*}

Figure \ref{fig:recency_outcome_correctness} assesses recency bias by splitting questions into older (Interval 1) and recent (Interval 2) subsets. The recency split uses a fixed midpoint at 2014-10-30: Interval 1 is $t < \text{2014-10-30}$ and Interval 2 is $t \ge \text{2014-10-30}$. This date is the range midpoint of the  Pre-Amendment and the Multi-Provision Pre-Amendment questions (min 2005-03-15, max 2024-06-15). Average performance improves slightly for recent questions (rising $\approx$ 0.04--0.06). However, significant performance divergences ($p < 0.05$) are observed exclusively in the Web condition, where recent questions score approximately 0.20--0.24 points higher. This suggests that recency advantages are driven primarily by web retrieval characteristics rather than inherent biases in the models' parametric knowledge or the curated RAG pipelines.

\subsection{LLM Judge Validation}

To validate the LLM judge, we sampled 48 questions stratified by question type and legal domain, yielding 192 model answers, which a legal expert independently scored on the same four criteria. Overall agreement was strong ($r = 0.853$, $\kappa = 0.751$, MAE $= 0.123$) with negligible bias. Agreement was highest for outcome correctness ($\kappa = 0.860$), consistent with its near-binary nature, and lowest for legal basis correctness ($\kappa = 0.549$), where evaluating provision citations involves more subjective judgment. By question type, Post-Cutoff Amendment questions showed the highest agreement ($\kappa = 0.821$) and Pre-Amendment questions the lowest ($\kappa = 0.696$). RAG-based settings yielded higher judge--expert agreement ($\kappa = 0.733$--$0.805$) than Vanilla and Web ($\kappa = 0.593$--$0.639$). These results confirm that the LLM judge is suitable for large-scale evaluation.

\section{Discussion}

Our findings underscore a central tension in the use of LLMs for legal research: the static nature of model training is structurally misaligned with the dynamic nature of statutory law. This mismatch is practically salient given that a large share of legal professionals already report using AI systems for research tasks. 

\subsection{Knowledge Cutoff as a Critical Failure Point}

In this context, Post-Cutoff Amendment Question answering reveals a core vulnerability: without external grounding or tool-calling, even state-of-the-art models degrade substantially when asked to answer questions that depend on statutory amendments enacted after their training cutoff. It is important to qualify the performance metrics in this Vanilla setting. While models like ChatGPT-5.1 achieved an outcome correctness score of approximately 40\%, this figure masks a deeper failure. This partial success stems from the binary nature of many legal questions (e.g., "lawful" vs. "unlawful"), where a random guess yields a 50\% baseline accuracy. The discrepancy becomes evident when analyzing the Legal Reasoning Correctness, which dropped to 0\%. Crucially, the models rarely signaled uncertainty, as shown by the small number of refusals. In most cases, the models confidently applied superseded provisions, a behavior that poses significant liability risks in legal practice.

\subsection{RAG as a Valid Mitigation Strategy}

Across all three question types, retrieval-augmented approaches provide a robust mitigation. Both RAG-kNN and RAG-ToC consistently outperformed all baselines across all metrics. By extracting an "as-of" date from the query and strictly filtering the retrieval corpus to that validity period, the system forces the model to reason within the correct legal regime. This also implies that existing legal RAG systems should not simply retrieve "relevant" semantic matches but must also take into account the temporal validity of a statute as a hard constraint. The lack of significant difference between RAG-kNN and RAG-ToC suggests that what is retrieved (the correct timeframe) matters more than how it is retrieved (embeddings vs. table of contents). While RAG solved the retrieval problem, the Multi-Hop task demanded when answering Multi-Provision Pre-Amendment Questions highlighted the reasoning ceiling. Even with perfect retrieval, performance dropped slightly when models had to synthesize multiple historical provisions. However, the gap was modest, indicating that once the correct temporal context is provided, current LLMs are generally capable of the necessary legal reasoning.

To isolate the impact of the regex-based date extraction, we conducted a ablation on a subset of our questions. Removing the temporal filter reduced the system to a standard RAG pipeline, which caused a substantial drop in accuracy.

\subsection{Web Search Tool Calling Prone to Recency Bias}

Finally, our results caution against treating generic web search as a reliable remedy for temporal law application errors. While web-enabled configurations improve over LLM (vanilla) in many cases, performance is unstable and exhibits a systematic recency skew in historically anchored tasks. In our (Multi-Provision) Pre-Amendment experiments, LLM-web performed substantially better on recent questions than on older ones, consistent with retrieval pipelines that favor "fresh" content and thereby over-retrieve current statute versions even when queries concern historical regimes. This implies that browsing alone is not an adequate safeguard for temporally sensitive legal questions. Instead, systems should rely on curated, versioned statutes and explicit temporal filtering to ensure that the retrieved law matches the relevant point in time.

\section{Conclusion and Open Questions}
Our findings show that legal LLM-based systems that does not explicitly model temporal validity comes with reliability risks for real-world legal research, several important questions remain unresolved. While our RAG strategies successfully retrieve temporally valid provisions, Multi-Provision questions revealed a reasoning ceiling when models must synthesize multiple historical provisions that changed asynchronously. It remains to be investigated how our findings generalize to common law jurisdictions, where legal change is driven not only by legislative amendment but depends on characteristic dynamics around stare decisis and argumentation with precedent cases. These may influence how models learn and unlearn legislative history during training.

In addition, our evidence for recency bias is mixed. Although we observe a recency advantage and statistically significant performance divergences in the Web condition, the other comparisons do not yield statistically significant effects. This limits the extent to which we can claim a robust, model-intrinsic recency bias across settings. Establishing when and why recency bias emerges therefore remains an important direction for future research, ideally using larger samples.

\section{AI Use Disclosure}
Generative AI tools were used to produce parts of this work, including tables, figures and code, as well as to perform language corrections and stylistic refinements. 

\section{Acknowledgements}
We thank Elly Breu and Quirin Pelikan (LMU Munich) for helping annotate the QA dataset. We also thank the \texttt{buzer.de} project for its very valuable open-access data contribution to the German legal community. 
This work was funded by the Digitalisierungsinitiative des Bundes für die Justiz under the "Generatives Sprachmodell der Justiz (GSJ)" project, a collaboration between the justice ministries of North Rhine-Westphalia and Bavaria, TU Munich, and the University of Cologne, and by the German Federal Ministry of Finance under the "Machbarkeitsstudie Digitaltaugliches Steuerrecht" project, a collaboration between LMUDigiTax and TU Munich.

\bibliographystyle{ACM-Reference-Format}
\bibliography{acmart}

@inproceedings{NEURIPS2021_f5bf0ba0,
 author = {Lazaridou, Angeliki and Kuncoro, Adhi and Gribovskaya, Elena and Agrawal, Devang and Liska, Adam and Terzi, Tayfun and Gimenez, Mai and de Masson d\textquotesingle Autume, Cyprien and Kocisky, Tomas and Ruder, Sebastian and Yogatama, Dani and Cao, Kris and Young, Susannah and Blunsom, Phil},
 booktitle = {Advances in Neural Information Processing Systems},
 pages = {29348--29363},
 publisher = {Curran Associates, Inc.},
 title = {Mind the Gap: Assessing Temporal Generalization in Neural Language Models},
 url = {https://proceedings.neurips.cc/paper_files/paper/2021/file/f5bf0ba0a17ef18f9607774722f5698c-Paper.pdf},
 volume = {34},
 year = {2021}
}

@inproceedings{chenghaozhu-etal-2025-llm,
   title={Is Your LLM Outdated? A Deep Look at Temporal Generalization},
   url={http://dx.doi.org/10.18653/v1/2025.naacl-long.381},
   DOI={10.18653/v1/2025.naacl-long.381},
   booktitle={Proceedings of the 2025 Conference of the Nations of the Americas Chapter of the Association for Computational Linguistics: Human Language Technologies (Volume 1: Long Papers)},
   publisher={Association for Computational Linguistics},
   author={Zhu, Chenghao and Chen, Nuo and Gao, Yufei and Zhang, Yunyi and Tiwari, Prayag and Wang, Benyou},
   year={2025},
   pages={7433–7457} }

@inproceedings{t-y-s-s-etal-2024-chronoslex,
    title = "{C}hronos{L}ex: Time-aware Incremental Training for Temporal Generalization of Legal Classification Tasks",
    author = "T.y.s.s, Santosh  and
      Vuong, Tuan-Quang  and
      Grabmair, Matthias",
    booktitle = "Proceedings of the 62nd Annual Meeting of the Association for Computational Linguistics (Volume 1: Long Papers)",
    month = aug,
    year = "2024",
    url = "https://aclanthology.org/2024.acl-long.166/",
    doi = "10.18653/v1/2024.acl-long.166",
    pages = "3022--3039"
}

@inproceedings{t-y-s-s-vuong-2025-lextempus,
    title = "{L}ex{T}empus: Enhancing Temporal Generalizability of Legal Language Models Through Dynamic Mixture of Experts",
    author = "T.y.s.s, Santosh  and
      Vuong, Tuan-Quang",
    booktitle = "Proceedings of the 63rd Annual Meeting of the Association for Computational Linguistics (Volume 1: Long Papers)",
    month = jul,
    year = "2025",
    url = "https://aclanthology.org/2025.acl-long.329/",
    doi = "10.18653/v1/2025.acl-long.329",
    pages = "6608--6624",
    ISBN = "979-8-89176-251-0"
}

@inproceedings{NEURIPS2020_6b493230,
 author = {Lewis, Patrick and Perez, Ethan and Piktus, Aleksandra and Petroni, Fabio and Karpukhin, Vladimir and Goyal, Naman and K\"{u}ttler, Heinrich and Lewis, Mike and Yih, Wen-tau and Rockt\"{a}schel, Tim and Riedel, Sebastian and Kiela, Douwe},
 booktitle = {Advances in Neural Information Processing Systems},
 pages = {9459--9474},
 publisher = {Curran Associates, Inc.},
 title = {Retrieval-Augmented Generation for Knowledge-Intensive NLP Tasks},
 url = {https://proceedings.neurips.cc/paper_files/paper/2020/file/6b493230205f780e1bc26945df7481e5-Paper.pdf},
 volume = {33},
 year = {2020}
}

@misc{zheng2023judging,
  title        = {Judging {LLM}-as-a-Judge with {MT}-Bench and Chatbot Arena},
  author       = {Zheng, Lianmin and Chiang, Wei-Lin and Sheng, Ying and Zhuang, Siyuan and Wu, Zhanghao and Zhuang, Yonghao and Lin, Zi and Li, Zhuohan and Li, Dacheng and Xing, Eric P. and Zhang, Hao and Gonzalez, Joseph E. and Stoica, Ion},
  year         = {2023},
  archivePrefix= {arXiv},
  doi          = {10.48550/arXiv.2306.05685}
}

@inproceedings{enguehard2025lemaj,
  title     = {{L}e{MAJ} (Legal {LLM}-as-a-Judge): Bridging Legal Reasoning and {LLM} Evaluation},
  author    = {Enguehard, Joseph and Van Ermengem, Morgane and Atkinson, Kate and Cha, Sujeong and Chowdhury, Arijit Ghosh and Ramaswamy, Prashanth Kallur and Roghair, Jeremy and Marlowe, Hannah R. and Negreanu, Carina Suzana and Boxall, Kitty and Mincu, Diana},
  booktitle = {Proceedings of the Natural Legal Language Processing Workshop 2025},
  year      = {2025},
  pages     = {318--337},
  url       = {https://aclanthology.org/2025.nllp-1.23/},
  doi       = {10.18653/v1/2025.nllp-1.23}
}

@inproceedings{10.1145/3709025.3712219,
author = {Zheng, Lucia and Guha, Neel and Arifov, Javokhir and Zhang, Sarah and Skreta, Michal and Manning, Christopher D. and Henderson, Peter and Ho, Daniel E.},
title = {A Reasoning-Focused Legal Retrieval Benchmark},
year = {2025},
isbn = {9798400714214},
publisher = {Association for Computing Machinery},
url = {https://doi.org/10.1145/3709025.3712219},
doi = {10.1145/3709025.3712219},
booktitle = {Proceedings of the 2025 Symposium on Computer Science and Law},
pages = {169–193},
}

@inproceedings{vladika-etal-2025-facts,
    title = "Facts Fade Fast: Evaluating Memorization of Outdated Medical Knowledge in Large Language Models",
    author = "Vladika, Juraj  and
      Dhaini, Mahdi  and
      Matthes, Florian",
    booktitle = "Findings of the Association for Computational Linguistics: EMNLP 2025",
    month = nov,
    year = "2025",
    url = "https://aclanthology.org/2025.findings-emnlp.487/",
    doi = "10.18653/v1/2025.findings-emnlp.487",
    pages = "9161--9174",
    ISBN = "979-8-89176-335-7"
}

@misc{buzer2026,
  author       = {{buzer.de}},
  title        = {Online Database of German Laws},
  howpublished = {\url{https://www.buzer.de}},
  year         = {2026},
  note         = {Accessed December 16, 2025}
}

@inproceedings{prior-etal-2025-risks,
    title = "Risks and Limits of Automatic Consolidation of Statutes",
    author = "Prior, Max  and
      Hof, Adrian  and
      Wais, Niklas  and
      Grabmair, Matthias",
    booktitle = "Proceedings of the Natural Legal Language Processing Workshop 2025",
    month = nov,
    year = "2025",
    publisher = "Association for Computational Linguistics",
    url = "https://aclanthology.org/2025.nllp-1.29/",
    doi = "10.18653/v1/2025.nllp-1.29",
    pages = "396--407",
    ISBN = "979-8-89176-338-8"
}

@article{wen-etal-2025-know,
    title = "Know Your Limits: A Survey of Abstention in Large Language Models",
    author = "Wen, Bingbing  and
      Yao, Jihan  and
      Feng, Shangbin  and
      Xu, Chenjun  and
      Tsvetkov, Yulia  and
      Howe, Bill  and
      Wang, Lucy Lu",
    journal = "Transactions of the Association for Computational Linguistics",
    volume = "13",
    year = "2025",
    publisher = "MIT Press",
    url = "https://aclanthology.org/2025.tacl-1.26/",
    doi = "10.1162/tacl_a_00754",
    pages = "529--556"
}

@misc{surani2025aiscalinglegalreform,
      title={AI for Scaling Legal Reform: Mapping and Redacting Racial Covenants in Santa Clara County}, 
      author={Faiz Surani and Mirac Suzgun and Vyoma Raman and Christopher D. Manning and Peter Henderson and Daniel E. Ho},
      year={2025},
      archivePrefix={arXiv},
      url={https://arxiv.org/abs/2503.03888}, 
}

@article{zhang2025llms_overruled_precedent,
  title        = {Do LLMs Truly Understand When a Precedent Is Overruled?},
  author       = {Li Zhang and Jaromír Savelka and Kevin Ashley},
  year         = {2025},
  archivePrefix= {arXiv},
  url          = {https://arxiv.org/abs/2510.20941}
}

@inproceedings{10.1145/3767695.3769493,
author = {Fang, Hanpei and Tao, Sijie and Chen, Nuo and Chang, Kai-Xin and Sakai, Tetsuya},
title = {Do Large Language Models Favor Recent Content? A Study on Recency Bias in LLM-Based Reranking},
year = {2025},
isbn = {9798400722189},
publisher = {Association for Computing Machinery},
url = {https://doi.org/10.1145/3767695.3769493},
doi = {10.1145/3767695.3769493},
booktitle = {Proceedings of the 2025 Annual International ACM SIGIR Conference on Research and Development in Information Retrieval in the Asia Pacific Region},
pages = {85–94},
numpages = {10},
}

@inproceedings{buttner-habernal-2024-answering,
  title     = "Answering legal questions from laymen in {G}erman civil law system",
  author    = {B{\"u}ttner, Marius and Habernal, Ivan},
  booktitle = "Proceedings of the 18th Conference of the European Chapter of the Association for Computational Linguistics (Volume 1: Long Papers)",
  month     = mar,
  year      = "2024",
  url       = "https://aclanthology.org/2024.eacl-long.122/",
  doi       = "10.18653/v1/2024.eacl-long.122",
  pages     = "2015--2027"
}

@misc{fan2025lexambenchmarkinglegalreasoning,
      title={LEXam: Benchmarking Legal Reasoning on 340 Law Exams}, 
      author={Yu Fan and Jingwei Ni and Jakob Merane and Yang Tian and Yoan Hermstrüwer and Yinya Huang and Mubashara Akhtar and Etienne Salimbeni and Florian Geering and Oliver Dreyer and Daniel Brunner and Markus Leippold and Mrinmaya Sachan and Alexander Stremitzer and Christoph Engel and Elliott Ash and Joel Niklaus},
      year={2025},
      archivePrefix={arXiv},
      url={https://arxiv.org/abs/2505.12864}, 
}

@techreport{thomsonreuters2025genai_profservices,
  title       = {Generative AI in Professional Services Report},
  author      = {Thomson Reuters},
  year        = {2025},
  url         = {https://www.thomsonreuters.com/content/dam/ewp-m/documents/thomsonreuters/en/pdf/reports/2025-generative-ai-in-professional-services-report-tr5433489-rgb.pdf},
}

@article{10.1093/biomet/34.1-2.28,
  title={The generalization of 'Student's' problem when several different population variances are involved},
  author={Welch, Bernard L},
  journal={Biometrika},
  volume={34},
  number={1/2},
  pages={28--35},
  year={1947},    
  doi = {10.1093/biomet/34.1-2.28},
  url = {https://doi.org/10.1093/biomet/34.1-2.28},
  publisher={JSTOR}
}

@techreport{bav2024ki_umfrage,
  title        = {Auswertung der Umfrage zur KI-Nutzung der bayerischen Anwaltschaft},
  author       = {Bayerischer Anwaltverband e.V.},
  year         = {2024},
  institution  = {Bayerischer Anwaltverband e.V.},
  url          = {https://www.bayerischer-anwaltverband.de/site/assets/files/1765/umfrage_zur_ki_nutzung_der_bay_anwaltschaft.pdf}
}

\end{document}